\documentclass[12pt]{article}
\usepackage[T1]{fontenc}

\usepackage{fancyhdr,lipsum}
\usepackage{graphicx} 
\usepackage[style=numeric,backend=biber,sorting=none]{biblatex}
\usepackage{float}
\usepackage[hyperindex,breaklinks,citecolor = blue]{hyperref}
\usepackage{authblk}
\usepackage[hyphenbreaks]{breakurl}
\usepackage{url}

\hypersetup{
    breaklinks=true,   
    colorlinks=true,   
    pdfusetitle=false,  
    citecolor=red,
    colorlinks=true}

\title{Using Letter Positional Probabilities to Assess Word Complexity}
\author{Michael C. Dalvean$^{}$  \\
        \small $^{}$School of Computer Science and Software Engineering \\
         \small$^{}$University of Western Australia\\
         \small$^{}$35 Stirling Highway, Perth WA 6009\\
         \small$^{}$michael.dalvean@research.uwa.edu.au\\
         \small$^{}$ORCID: 0000-0003-3689-1258\\
}

\begin{document}

\date{August 2024}

\maketitle

\begin{abstract}
 Word complexity is defined in a number of different ways. Psycholinguistic, morphological and lexical proxies are often used. Human ratings are also used. The problem here is that these proxies do not measure complexity directly, and human ratings are susceptible to subjective bias. In this study we contend that some form of `latent complexity' can be approximated by using samples of simple and complex words.  We use a sample of `simple' words from primary school picture books and a sample of `complex' words from high school and academic settings. In order to analyse the differences between these groups, we look at the letter positional probabilities (LPPs). We find strong statistical associations between several LPPs and complexity. For example, simple words are significantly ($p<.001$) more likely to start with w, b, s, h, g, k, j, t, y or f, while complex words are significantly ($p<.001$) more likely to start with i, a, e, r, v, u or d. We find similar strong associations for subsequent letter positions, with 84 letter-position variables in the first 6 positions being significant at the $p<.001$ level. We then use LPPs as variables in creating a classifier which can classify the two classes with an 83\% accuracy. We test these findings using a second data set, with 66 LPPs significant ($p<.001$) in the first 6 positions common to both datasets. We use these 66 variables to create a classifier that is able to classify a third dataset with an accuracy of 70\%.  Finally, we create a fourth sample by combining the extreme high and low scoring words generated by three classifiers built on the first three separate datasets and use this sample to build a classifier which has an accuracy of 97\%. We use this to score the four levels of English word groups from an ESL program. 
\end{abstract}


\section{Introduction}

The issue of text and word complexity has gained significant interest in recent years. Text simplification, text complexity assessment and development of educational resources are direct applications of word complexity analysis. The purpose of this paper is to examine the relationship between the basic ‘atomic’ structure of words and `complexity’. 

In this paper, complexity is taken to mean an overall measure of the difference between the kinds of words used in primary school picture books and the kinds of words used in high school textbooks and beyond. There are a number of proxies we can use for word complexity, such as age of acquisition, frequency in written and spoken English, and psycholinguistic measures such as familiarity and concreteness. However, our initial examination begins with a sample of primary-school words and a sample of non-primary-school words. We use these other measures of word complexity to create datasets of complex/simple words to test the findings from the initial dataset. We then create a dataset similar to the first to test the common significant variables in the first two datasets. The overall approach is based on using samples of identified simple and complex words which, we propose, represent two clusters on a spectrum representing latent complexity (LC).  

We look at the ‘atomic’ structure of words in this paper. Atomic is used here to denote individual letter placement in the word. More ‘molecular’ methods have been used in previous work. For example, word endings have been used to denote abstraction \parencite{rabinovich2018learning}, which is positively associated with word complexity. Word lengths, syllable counts and phonotactic measures have also been used to define word complexity. However, individual letter positional probabilities have not, to the author’s knowledge, been used as a basis for determining complexity, although there has been some word done on the effects of perception of individual letters on reading \parencite{bouwhuis1979visual}, and there has been work on the effects on word shape \parencite{larson2004science}. We find that there are many letter positional probabilities that are associated with word complexity. For example, we find that high LC words are significantly ($p<.01$) more likely to start with a vowel than low LC words.

One of the benefits of using letter positional probabilities is that it gives us good insight into the relationship between word complexity and the low-level construction of words. In educational settings, it has been found that reading and phonological speech errors occur at a significantly higher rate in relation to vowels than consonants \parencite{post1999identification}. By using letter positional probabilities to analyse words, we can get some insight into this phenomenon in that we can demonstrate how different combinations of vowels and consonants work to influence the overall complexity of a word. This enables us to rank individual words on LC at a fine-grained level. 

\section{Previous Work}

The standard approach to word complexity prediction is to start with a ‘gold standard’ of word complexity according to human assessors. This can be in the form of a binary assessment - complex or not complex -, or some kind of graduated scale \parencite{shardlow2022predicting}. These are then used as the dependent variables in regression models. The independent variables are such measures as structural, psycholinguistic lexical characteristics of words. The problem with human assessments is that they are essentially subjective. For example, human raters give more complex objects longer novel names in experimental settings \parencite{lewis2016length}. Human ratings are also likely to be influenced by language competence \parencite{gooding2022one}. 

Word length has been found to be positively associated with complexity \parencite{dale1948formula,mc1969smog,kearns2022word}, although there is some evidence that frequency is more important than word length \parencite{wilkens2014size}. Closely related to word length is number of syllables. Given that length and complexity are associated, we would expect that syllable number would be associated with complexity. However, controlling for length, human coders pair more spatially complex objects with names with greater numbers of syllables than simpler objects \parencite{kelly1990relation}, indicating that syllable count is more important than length. 

Age of acquisition (AoA) has been found to be closely related to phonological complexity \parencite{gendler2021effect} and word length \parencite{zevin2002age}. In general, words acquired earlier in life are processed more quickly than those acquired later \parencite{hernandez2007age}. The link with complexity is that less complex words tend to be acquired at earlier ages. This is why AoA is seen as as a good proxy for complexity.

Frequency is a proxy for several factors associated with complexity  \parencite{Kucera1967,rayner1986lexical}. Frequency is usually assessed on the basis of large written or spoken corpora by calculating how often the word appears in a given corpus divided by the number of words in the corpus. The link with complexity is that words that are used frequently are less complex than those that are used infrequently. The main issue with using frequency as a proxy for complexity is that actual frequency is associated with a given national context \parencite{brysbaert2009moving}, and this means that the complexity associated with frequency may be dependent on the region of application.

Closely related to frequency is prevalence, which is a measure of familiarity/frequency derived from the number of annotators who know a given word divided by the number of annotators \parencite{keuleers2015word}. Brysbaert et al \parencite{brysbaert2016impact} found in a Dutch setting that Prevalence was a better predictor of response time in lexical decision making than frequency. 

Concreteness is linked to complexity in that more concrete words tend to be less conceptually complex \parencite{hiebert2019analysis}. Concreteness is a measure of the extent to which a word can be associated with a sensory perception. Concrete words are easier to recall \parencite{walker1999concrete}, easier to read and classify \parencite{sysoeva2007rapid}, easier to learn \parencite{mestres2014mapping}, and are learnt by recruitment of different brain areas in comparison with abstract words \parencite{mestres2009functional}. The traditional way of measuring concreteness is to ask human raters to assess the extent to which a word is easy to associate with a physical phenomenon. Other methods include assessing the nature of the word. For example, words ending in ‘-ling’ (eg: earthling) denote an individual (relatively concrete), while ‘-hood’ (eg: childhood) denotes a condition (relatively abstract) \parencite{rabinovich2018learning}. Buccino et al \parencite{buccino2019concreteness} propose that concreteness is associated with complexity in that concrete words are linked to less complex experience than abstract words. That is, while both types of words are linked to experience, the experiences associated with abstract words are more complex than experiences associated with concrete words. 

Much of the recent work in complex word identification (CWI) has been designed to identify complex words for the purposes of text simplification. Gooding and Tragut \parencite{gooding2022one}, for example, find that a personalised approach to CWI is appropriate because there are likely to be idiosyncratic differences in perceptions of complexity. Similarly, Lee and Yeung \parencite{lee2019personalized} find that, with human coders,  candidate substitute simple words that are aligned with the coder’s English proficiency optimise the trade off between complexity and semantic faithfulness. North and Zampieri \parencite{north2023features} find that idiosyncratic assessment of word complexity extends to L1 and L2 language learners in that complexity based on 6 linguistic characteristics of words have a greater effect on perceived complexity on L1 than on L2 learners. 

\section{The Current Study}

We start from the point of view that complexity is a latent variable for which there are proxies such as those discussed above. In order to capture the quality of complexity in samples of words, we use two sets of words from settings that differ in terms of the level of complexity associated with each setting. Words commonly found in children’s picture books we would intuitively describe as simple in comparison to words used in high school and beyond. Certainly, we would find that samples from each setting would differ on the proxies we have mentioned above, but the underlying variable we are interested in is that instantiated in the differences between the two samples. For want of a better term, we describe the complexity we refer to as ‘latent complexity’ (LC).

Rather than use any particular proxies to measure the differences between the two samples, we use the ‘atomic’ structure of the words. We measure the atomic structure of each word by breaking words into binary values for each letter in the word. That is, based on a 26-letter alphabet, we ‘one-hot-code’  the letter in each position of a word up to word lengths of 22 letters. Thus, the first 26 variables consist of binary values for the first letter of the word, the second 26 variables consist of binary values for the second letter of the word, and so on up to 22 letters. In this way, each word is reduced to (22*26) binary variables. 

We demonstrate that the actual letter placement in words has a significant association with LC.  In Experiment 1 we find 84 letter positional probabilities that are highly ($p<.001$) associated with LC in the first 6 letter positions. In Experiment 2 we extend this analysis by creating a second corpus of simple and complex words. We create the class of simple words by combining samples of words that score high on frequency and concreteness, and low on Age of acquisition. Similarly, we create the complex word class by taking words from the opposite ends of the spectrum of these same measures. We find essentially the same pattern of positional letter probabilities in the expanded corpus. In all, we find 227 letter positional probabilities that are significantly ($p<.001$) associated with LC and 238 that are significant at the p$<.01$ level. Importantly, we find 66 variables that are significant at the p$< .001$ level that are common as between the two datsets. We use these 66 variables to create a classifier that is able to classify a third dataset with an accuracy of 70\%. Thus, we find that there are common LPPs that are predictive of LC across three datasets.

\subsection{Nomenclature}

In the following we report the p-values for variables in the experiments. Each experiment involves analysing a large number of variables. This is due to there being a large number of variables under examination as a result of creating 26 binary variables for each letter in words of length up to 18 letters in Experiment 1 and 22 letters in Experiments 2 and 3. With 22 letters there are 22* 26 = 572 binary variables.   Approximately 5 of these may be significant at the $p<.01$ by chance alone. For the purposes of assessing the p-values, we must account for the large number of independent t-tests we are performing. The method we use is Bonferroni correction. This method multiplies the p-value by the number of tests. Thus, in experiment 2, we multiply each p-value by 572. In experiment 1, with a maximum word length of 18, the relevant multiplier is 468. We denote the Bonferroni adjusted p-values as Bp.

Another issue is that we report the Bp to five decimal places. The reason for this is that we wish to emphasise the strength of the statistical associations we report in this study. In the tables that follow, a Bp-value of 0 represents a Bp of $<.00001$.

All of the independent variables in this paper are letter positional variables represented by the lower case letter and the position. Thus, `w1' represents the letter `w' in the first position.

\section{Experiment 1}

\subsection{Method}

In this experiment we create a dataset of complex and simple words. We then analyse the salient letter positional probabilities to determine which letters in which positions are important for distinguishing between the two classes. Finally, we create a classifier to distinguish between the two categories of words based on letter positional probabilities. 

\subsection{Data}

Several data sources were used for the experiments. For the first experiment, we use the Children’s Picture Books Lexicon \parencite{green2023children} to create the simple class. This is a corpus of 25,585 words from 2146 children’s picture books. Words that appeared in at least 3 texts were extracted, giving 10773 words. The complex word class was created from two sources: 1) a list of 2925 words commonly appearing in GRE tests was extracted from \parencite{Otswald_gre}; 2) several word lists were extracted from \parencite{academic_lists}, as shown in Table \ref{tab:lists} 

Words containing hyphens, apostrophes or any other non-alphabetical characters were excluded. Single letters were excluded. Note that the two letter words included standard two letter words as well as forms such as  ‘eh’, ah’ , ‘aw’ etc. Acronyms were also included.

\begin{center}

\begin{table}[H]
\centering
\caption{Academic Word Lists}
\label{tab:lists}
\begin{tabular}{|l|l|}
\hline 
List & No. Words\tabularnewline\hline
\hline 
Computer Science Word List & 433\tabularnewline
\hline 
Academic Vocabulary for Middle School Students & 1923\tabularnewline
\hline 
Engineering Technical Word List & 313\tabularnewline
\hline 
Medical Academic Word List & 623\tabularnewline
\hline 
Secondary Vocabulary List & 4780\tabularnewline
\hline 
University Word List & 806\tabularnewline
\hline 
New Academic Word List & 2598\tabularnewline
\hline 
General Service List & 2949\tabularnewline
\hline 
Academic Word List & 3109\tabularnewline
\hline 
Academic Vocabulary List & 175\tabularnewline
\hline 
\end{tabular}
\end{table}
\end{center}

Combining these sources results in a class of 13273 complex words. For both classes, Roman Numerals and words containing non-alphabetic characters were removed. Where a word was duplicated in a class the duplicate(s) were removed leaving only one example. This step is included because in some datasets the same word can occur as, for example, a noun and a verb with different scores. In samples where this is not the case, such as the current samples, the numbers for `Cleaned Words' and `Within Class Duplicates Removed' are the same. However, this step is included as a test for the possible existence of within-class duplicates. Words that were present in both classes, `Between Class Common Words' were removed. The rationale here was that if a word occurs in the complex and the simple class, there is no ground truth as to whether it is a simple or a complex word. After filtering we are left with a final class of 7574 simple words and 10212 complex words, as shown in Table \ref{tab:table01}.

\begin{center}
\begin{table}[H]
\centering
\caption{Word Filtering - Experiment 1}
\label{tab:table01}
\begin{tabular}{|l|l|l|}
\hline
                                 & Simple & Complex \tabularnewline\hline
Total Words                      & 10773  & 13273   \tabularnewline\hline
Cleaned Words                    & 10593  & 13231   \tabularnewline\hline
Within Class Duplicates Removed  & 10593  & 13231   \tabularnewline\hline
Between Class Common Words Removed & 7574   & 10212  \tabularnewline\hline
\end{tabular}
\end{table}
\end{center}

\subsection{Variables}

The method of analysis entails breaking words down into binary representations of their constituent letter positions. The longest word in the corpus in the first experiment is 18 letters. Thus, there are 18 * 26 = 468 binary variables, with the first 26 variables representing the first letter, the second 26 variables representing the second letter, and so on. Thus, the word ‘cat’ has a binary value of 1 for the letter ‘c’ in the first 26 variables, with all 25 other variables taking a value of 0, and so on for ‘a’ in the second set of binary 26 variables and ‘t’ for the third set of 26 binary variables. All values for the remaining sets of variables, representing letter positions 4 – 18 are set at 0.

The variables generated by the above process are then used to calculate p-values for t-tests for each variable. Thus, for the letter `a' in first position (a1) we test for a significant difference between the probability of the letter `a' being in first position in the simple word sample as opposed to the complex word sample. The `Letter Positional Probabilities' in the following tables are calculated by taking the number of times a given letter occurs in a given position and dividing by the sample size. Thus, we see in Table \ref{tab:table02} in the fourth row that the letter `a' occurs in first position at a rate of 0.37 in the simple sample and 0.88 in the complex sample size.  

\subsection {Results}

There are 468 variables under analysis. Therefore we need to make a Bonferroni correction for 468 p-values. Thus, we multiply p-values by 468 to get the Bonferroni p. 

There are 195 variables that are significant at the Bp $<.01$ level and 183 that are significant at Bp$<.001$. However, many of these are due to the fact that the average word length of complex words is greater than that for simple words. As such, variables representing higher letter positions capture the word-length effect rather than the complexity effect. To reduce the length effect, we focus on variables within the range of both complex and simple words. The average length of simple words is 6.4 letters and 8.8 for complex words. To focus on the common variables we isolate the variables in the first 6 positions of the words. Of these, we focus on the variables that are significant at the Bp$<.001$ level.

There are 84 variables that are significant at the Bp$<.001$ level in the first 6 letter positions. Table \ref{tab:table02} details the 84 significant variables. The table is ordered by ascending position followed by ascending Bp level

\begin{center}
\begin{table}[H]
\footnotesize
\centering
\caption{Letter Positional Probabilities and Bp scores - Experiment 1}
\label{tab:table02}
\begin{tabular}{|p{.5in}|p{.5in}|p{.5in}|p{.5in}|p{.5in}|p{.5in}|p{.5in}|p{.5in}|}
\hline
Variable & Simple& Complex& BP      & Variable & Simple& Complex& BP      \tabularnewline\hline
i1       & 0.013          & 0.081           & 0       & s3       & 0.052          & 0.072           & 0.00003 \tabularnewline\hline
w1       & 0.045          & 0.008           & 0       & m3       & 0.036          & 0.052           & 0.00038 \tabularnewline\hline
b1       & 0.079          & 0.027           & 0       & k4       & 0.04           & 0.008           & 0       \tabularnewline\hline
a1       & 0.037          & 0.088           & 0       & i4       & 0.058          & 0.114           & 0       \tabularnewline\hline
s1       & 0.153          & 0.089           & 0       & u4       & 0.018          & 0.049           & 0       \tabularnewline\hline
e1       & 0.024          & 0.063           & 0       & w4       & 0.018          & 0.003           & 0       \tabularnewline\hline
h1       & 0.049          & 0.023           & 0       & o4       & 0.041          & 0.073           & 0       \tabularnewline\hline
g1       & 0.045          & 0.022           & 0       & g4       & 0.036          & 0.021           & 0       \tabularnewline\hline
k1       & 0.013          & 0.002           & 0       & d4       & 0.047          & 0.03            & 0       \tabularnewline\hline
j1       & 0.017          & 0.005           & 0       & z4       & 0.007          & 0.002           & 0.00003 \tabularnewline\hline
t1       & 0.066          & 0.04            & 0       & n4       & 0.066          & 0.047           & 0.00003 \tabularnewline\hline
r1       & 0.042          & 0.067           & 0       & y4       & 0.012          & 0.005           & 0.00091 \tabularnewline\hline
y1       & 0.007          & 0.001           & 0       & r5       & 0.044          & 0.102           & 0       \tabularnewline\hline
v1       & 0.009          & 0.021           & 0       & k5       & 0.032          & 0.005           & 0       \tabularnewline\hline
u1       & 0.011          & 0.022           & 0       & u5       & 0.014          & 0.043           & 0       \tabularnewline\hline
d1       & 0.055          & 0.074           & 0.00016 & o5       & 0.037          & 0.076           & 0       \tabularnewline\hline
f1       & 0.048          & 0.034           & 0.00077 & y5       & 0.032          & 0.011           & 0       \tabularnewline\hline
n2       & 0.037          & 0.1             & 0       & i5       & 0.075          & 0.119           & 0       \tabularnewline\hline
h2       & 0.061          & 0.025           & 0       & a5       & 0.039          & 0.071           & 0       \tabularnewline\hline
a2       & 0.159          & 0.099           & 0       & c5       & 0.019          & 0.043           & 0       \tabularnewline\hline
w2       & 0.016          & 0.001           & 0       & h5       & 0.033          & 0.017           & 0       \tabularnewline\hline
e2       & 0.115          & 0.171           & 0       & e5       & 0.143          & 0.109           & 0       \tabularnewline\hline
x2       & 0.005          & 0.02            & 0       & m5       & 0.017          & 0.031           & 0       \tabularnewline\hline
l2       & 0.064          & 0.038           & 0       & v5       & 0.003          & 0.011           & 0       \tabularnewline\hline
s2       & 0.005          & 0.016           & 0       & s5       & 0.08           & 0.057           & 0       \tabularnewline\hline
y2       & 0.006          & 0.018           & 0       & p5       & 0.028          & 0.017           & 0.00009 \tabularnewline\hline
d2       & 0.004          & 0.013           & 0       & n5       & 0.044          & 0.061           & 0.00013 \tabularnewline\hline
b2       & 0.003          & 0.011           & 0       & a6       & 0.026          & 0.086           & 0       \tabularnewline\hline
m2       & 0.011          & 0.023           & 0       & t6       & 0.034          & 0.093           & 0       \tabularnewline\hline
i3       & 0.084          & 0.049           & 0       & i6       & 0.056          & 0.121           & 0       \tabularnewline\hline
t3       & 0.049          & 0.084           & 0       & c6       & 0.013          & 0.04            & 0       \tabularnewline\hline
u3       & 0.055          & 0.029           & 0       & u6       & 0.008          & 0.029           & 0       \tabularnewline\hline
o3       & 0.085          & 0.055           & 0       & m6       & 0.008          & 0.028           & 0       \tabularnewline\hline
a3       & 0.113          & 0.08            & 0       & e6       & 0.132          & 0.088           & 0       \tabularnewline\hline
c3       & 0.038          & 0.063           & 0       & l6       & 0.03           & 0.051           & 0       \tabularnewline\hline
p3       & 0.029          & 0.049           & 0       & d6       & 0.048          & 0.029           & 0       \tabularnewline\hline
w3       & 0.012          & 0.004           & 0       & y6       & 0.026          & 0.012           & 0       \tabularnewline\hline
k3       & 0.009          & 0.003           & 0       & p6       & 0.007          & 0.018           & 0       \tabularnewline\hline
v3       & 0.012          & 0.024           & 0.00001 & k6       & 0.006          & 0.001           & 0       \tabularnewline\hline
e3       & 0.077          & 0.056           & 0.00002 & o6       & 0.024          & 0.041           & 0       \tabularnewline\hline
z3       & 0.007          & 0.002           & 0.00002 & x6       & 0              & 0.004           & 0.00001 \tabularnewline\hline
f3       & 0.012          & 0.023           & 0.00003 &          &                &                 &  
\tabularnewline\hline
\end{tabular}
\end{table}
\end{center}

An initial observation we can make about the results is that the probabilities for the occurrence of a, e, i and u in first position are significantly higher for complex words than for simple words. Thus, complex words are characterised by having vowels in first position, while simple words are characterised by having consonants in first position. A similar pattern exists for positions 4, 5 and 6.  

\subsection{Ranking Words by Complexity}

Given the large number of variables that are able to characterise the difference between the two classes, we should be able to rank each word according to its position in relation to the simple set of words and the complex set of words. That is, we can place each word on a spectrum representing the general level of LC based on the two classes of words. We do this by using machine learning to create a classifier which assigns each word a score based on the extent to which it resembles the set of simple as opposed to the set of complex words.

The features used for the classifier are those that we have used for the above analysis. This gives us 468 binary variables. However, some of these variables have no data. For example, the letter j does not occur in the positions 7 – 15 for either class in the dataset. As such, these variables are removed. Removal of the null variables leaves us with 377 binary variables. The target variable is a binary value of 1 if the word is in the complex class and 0 if it is in the simple class.  The induction method we use is the Python sklearn random forest with no. estimators=100, max depth=None, and min samples split=5. Due to the imbalanced dataset, the imbalancelearn SMOTE method was used to balance the training data. We use 10-fold cross validation, and the scores for each word are based on the held out fold. Classification of results is based on a score cutpoint of 0.5, such that words scoring $<0.5$ were classed as ‘simple’ and those $>= 0.5$ were classed at ‘complex’.

The accuracy of the classifier is 83\%, (\,baseline = 51\% )\, with 85\% sensitivity and 79\% specificity. The Kappa value is 0.6433 (z = 85.8, p $<.0001$ – two tailed test). Thus, the classifier is able to efficiently classify the dataset. This indicates that the score given to each word is a reasonable estimate of the word’s complexity.

In order to get an insight into the nature of the difference between high and low LC words, we show high and low LC words for each word length category in Table \ref{tab:table03}. Note that Class in the table is the word class, with 0 = Simple and 1 = Complex, and the score is the LC generated by the classifier. 

\begin{center}
\begin{table}[H]

\centering
\caption{Lowest and Highest Scoring Words for each Word Length Category - Experiment 1}
\label{tab:table03}
\begin{tabular}{|l|l|l|l|l|}
\hline
Length & Word               & Class & Score & Syllables \tabularnewline\hline
2      & so                 & 0     & 0.02  & 1         \tabularnewline\hline
2      & ex                 & 1     & 0.7   & 1         \tabularnewline\hline
\textbf3      &\textbf {bee}                & \textbf0     &\textbf 0     &\textbf 1         \tabularnewline\hline
\textbf3      & \textbf{inn}                & \textbf{0}     & \textbf{0.69}  & \textbf{1}         \tabularnewline\hline
4      & bots               & 0     & 0     & 1         \tabularnewline\hline
4      & edit               & 1     & 0.82  & 2         \tabularnewline\hline
\textbf5      & \textbf{spook}              &\textbf {0}     &\textbf {0}     &\textbf {1}        \tabularnewline\hline
\textbf5      & \textbf{inter}              & {1}     & \textbf{0.97}  & \textbf{2}         \tabularnewline\hline
6      & crowns             & 0     & 0.01  & 1         \tabularnewline\hline
6      & impure             & 1     & 0.96  & 2         \tabularnewline\hline
\textbf{7}      & \textbf{stumped}            & \textbf{0}     & \textbf{0.01}  & \textbf{1}         \tabularnewline\hline
\textbf{7}      & \textbf{inhibit}            & \textbf{1}     & \textbf{0.99}  & \textbf{3}         \tabularnewline\hline
8      & watching           & 0     & 0.03  & 2         \tabularnewline\hline
8      & motivate           & 1     & 0.99  & 3         \tabularnewline\hline
\textbf{9}      & \textbf{scribbled}          &\textbf{0}     & \textbf{0.07}  & \textbf{2}         \tabularnewline\hline
\textbf{9}      & \textbf{affective}          & \textbf{1}     & \textbf{0.99}  & \textbf{3}         \tabularnewline\hline
10     & thankfully         & 0     & 0.09  & 3         \tabularnewline\hline
10     & liberalize         & 1     & 1     & 3         \tabularnewline\hline
\textbf{11}     & \textbf{christmassy}        & \textbf{0}     & \textbf{0.23} & \textbf{3}         \tabularnewline\hline
\textbf{11}     & \textbf{application}        & \textbf{1}     & \textbf{1}     & \textbf{3}         \tabularnewline\hline
12     & thoughtfully       & 0     & 0.18  & 3         \tabularnewline\hline
12     & constituting       & 1     & 1     & 4         \tabularnewline\hline
\textbf{13}     & \textbf{neighborhoods}      & \textbf{0}     & \textbf{0.19}  & \textbf{3}         \tabularnewline\hline
\textbf{13}     & \textbf{dramatization}      & \textbf{1}     & \textbf{1}     & \textbf{5}         \tabularnewline\hline
14     & disappointment     & 0     & 0.33  & 4         \tabularnewline\hline
14     & conceptualised     & 1     & 1     & 5         \tabularnewline\hline
\textbf{15}     & \textbf{kindergarteners}    & \textbf{0}     & \textbf{0.25}  & \textbf{4}         \tabularnewline\hline
\textbf{15}     & \textbf{industrialising}    & \textbf{1}     & \textbf{1}     & \textbf{6}         \tabularnewline\hline
16     & cyclophosphamide   & 1     & 0.66  & 5         \tabularnewline\hline
16     & differentiations   & 1     & 1     & 6         \tabularnewline\hline
\textbf{17}     & \textbf{counterproductive}  & \textbf{1}     & \textbf{0.59}  & \textbf{5}         \tabularnewline\hline
\textbf{17}     & \textbf{industrialisation}  & \textbf{1}     & \textbf{1}     & \textbf{7}         \tabularnewline\hline
18     & tetrachloromethane & 1     & 0.67  & 6         \tabularnewline\hline
18     & industrialisations & 1     & 1     & 7         \tabularnewline\hline

\end{tabular}
\end{table}
\end{center}

It is interesting to note that in 13 of the word pairs, the complex words have more syllables than simple words, and in 4 word pairs the number of syllables is equal. In none of the word pairs do the simple words have more syllables than complex words.

\section{Experiment 2}

\subsection{Data}

In order to test the generality of the above results, we create a synthetic dataset similar to that used in Experiment 1. We consider datasets consisting of words rated on factors that have been found to be associated with word complexity – Age of Acquisition, Concreteness, and Frequency. For each of these we select two classes at the extremes of the distributions to provide samples of highly simple and highly complex words. 

There are three sources of data for the analysis:

1) Age of Acquisition norms were taken from \parencite{brysbaert2017test}. The initial 44000 words used in the article has been updated by the authors of the article to create a list of 51715 words \parencite{AOA_brysbaert}. The lowest scoring 10000 words were used as the basis for the simple word list, while the highest scoring were used as the basis for the complex list.

2) Concreteness ratings for 37,058 words were taken from Brysbaert \parencite{CONC_brysbaert}. The top and bottom scoring 10000 words were used for the simple and complex list respectively.

3) Frequency ratings for 74286 words were taken from Brysbaert (2009 ) \parencite{freq_brysbaert}. 10000 of the high frequency words were used for the simple class. In the case of the low frequency words, there were 13902 words with the same score (=1) in alphabetical order. If we took the first 10000 words we would exclude a large sample of words from towards the end of the alphabet. As such, the 13902 words with a score of 1 were randomly selected to create a sample of 10000 words, which were added to the simple class.

These 30000-word classes of simple and complex words had words containing non-alphabetic characters removed. Removal of duplicates within each class and common words between classes led to a simple class of 18978 and a complex class of 24501, as shown in Table \ref{table:word_samp_30000}

\begin{center}
\begin{table}[H]
\centering
\caption{Word Filtering - Experiment 2}
\label{table:word_samp_30000}
\begin{tabular}{|l|l|l|}
\hline
                                 & Simple & Complex \tabularnewline\hline
Total Words                      & 30000  & 30000   \tabularnewline\hline
Cleaned Words                    & 29998  & 29957   \tabularnewline\hline
Within Class Duplicates Removed  & 21260  & 26783   \tabularnewline\hline
Between Class Common Words Removed & 18978  & 24501  \tabularnewline\hline
\end{tabular}
\end{table}
\end{center}

The maximum number of letters in the classes was 22, so Bonferroni correction is based on 26 * 22 = 572 variables under examination. There are 238 and 226 variables significant at the Bp $< .01$ and Bp $< .001$ level respectively.

Given the large number of significant variables, we focus those for the first 6 positions. This is less than the average word length of this dataset (7.04 letters), but enables us to compare the significant variables with those of the first experiment. There are 85 variables that are significant at the Bp $<.001$ level in the first 6 positions. Table \ref{tab:p-values_exp_2} shows the 85 variables significant at the Bp $<.001$ level. 

\begin{table}[H]
\footnotesize
\caption{Letter Positional Probabilities and Bp scores - Experiment 2}
\label{tab:p-values_exp_2}
\begin{tabular}{|p{.5in}|p{.5in}|p{.5in}|p{.5in}|p{.5in}|p{.5in}|p{.5in}|p{.5in}|}
\hline
Variable & Simple & Complex & Bp      & Variable & Simple & Complex & Bp      \tabularnewline\hline
u1       & 0.012          & 0.062           & 0       & u4       & 0.019          & 0.046           & 0       \tabularnewline\hline
i1       & 0.018          & 0.069           & 0       & i4       & 0.06           & 0.095           & 0       \tabularnewline\hline
s1       & 0.145          & 0.087           & 0       & o4       & 0.041          & 0.071           & 0       \tabularnewline\hline
b1       & 0.08           & 0.044           & 0       & a4       & 0.051          & 0.074           & 0       \tabularnewline\hline
w1       & 0.038          & 0.016           & 0       & r4       & 0.057          & 0.079           & 0       \tabularnewline\hline
e1       & 0.027          & 0.05            & 0       & w4       & 0.014          & 0.006           & 0       \tabularnewline\hline
a1       & 0.044          & 0.069           & 0       & d4       & 0.049          & 0.033           & 0       \tabularnewline\hline
d1       & 0.049          & 0.07            & 0       & x4       & 0              & 0.004           & 0       \tabularnewline\hline
t1       & 0.06           & 0.043           & 0       & s4       & 0.064          & 0.05            & 0       \tabularnewline\hline
h1       & 0.045          & 0.031           & 0       & n4       & 0.062          & 0.049           & 0       \tabularnewline\hline
j1       & 0.014          & 0.007           & 0       & t4       & 0.09           & 0.075           & 0.00004 \tabularnewline\hline
o1       & 0.016          & 0.025           & 0       & g4       & 0.032          & 0.024           & 0.00009 \tabularnewline\hline
l1       & 0.036          & 0.026           & 0       & l4       & 0.062          & 0.051           & 0.00019 \tabularnewline\hline
g1       & 0.037          & 0.027           & 0       & a5       & 0.047          & 0.08            & 0       \tabularnewline\hline
f1       & 0.05           & 0.039           & 0.00002 & e5       & 0.148          & 0.11            & 0       \tabularnewline\hline
n2       & 0.038          & 0.127           & 0       & k5       & 0.024          & 0.01            & 0       \tabularnewline\hline
a2       & 0.167          & 0.115           & 0       & r5       & 0.058          & 0.087           & 0       \tabularnewline\hline
h2       & 0.052          & 0.031           & 0       & u5       & 0.019          & 0.038           & 0       \tabularnewline\hline
m2       & 0.01           & 0.023           & 0       & o5       & 0.052          & 0.073           & 0       \tabularnewline\hline
x2       & 0.006          & 0.014           & 0       & m5       & 0.018          & 0.03            & 0       \tabularnewline\hline
v2       & 0.004          & 0.011           & 0       & s5       & 0.075          & 0.058           & 0       \tabularnewline\hline
y2       & 0.008          & 0.017           & 0       & y5       & 0.02           & 0.012           & 0       \tabularnewline\hline
w2       & 0.009          & 0.003           & 0       & i5       & 0.08           & 0.099           & 0       \tabularnewline\hline
t2       & 0.029          & 0.017           & 0       & c5       & 0.026          & 0.036           & 0       \tabularnewline\hline
o2       & 0.155          & 0.13            & 0       & w5       & 0.01           & 0.005           & 0       \tabularnewline\hline
l2       & 0.054          & 0.039           & 0       & n5       & 0.047          & 0.059           & 0.00001 \tabularnewline\hline
e2       & 0.129          & 0.155           & 0       & h5       & 0.037          & 0.028           & 0.00007 \tabularnewline\hline
b2       & 0.003          & 0.009           & 0       & g5       & 0.015          & 0.022           & 0.00021 \tabularnewline\hline
s2       & 0.006          & 0.013           & 0       & v5       & 0.005          & 0.009           & 0.00027 \tabularnewline\hline
i2       & 0.108          & 0.09            & 0       & i6       & 0.068          & 0.108           & 0       \tabularnewline\hline
u2       & 0.082          & 0.069           & 0.00012 & a6       & 0.047          & 0.075           & 0       \tabularnewline\hline
p3       & 0.032          & 0.052           & 0       & t6       & 0.048          & 0.075           & 0       \tabularnewline\hline
i3       & 0.071          & 0.049           & 0       & m6       & 0.012          & 0.028           & 0       \tabularnewline\hline
o3       & 0.081          & 0.057           & 0       & c6       & 0.022          & 0.04            & 0       \tabularnewline\hline
a3       & 0.104          & 0.079           & 0       & u6       & 0.013          & 0.028           & 0       \tabularnewline\hline
s3       & 0.055          & 0.077           & 0       & l6       & 0.035          & 0.055           & 0       \tabularnewline\hline
f3       & 0.013          & 0.022           & 0       & p6       & 0.01           & 0.022           & 0       \tabularnewline\hline
c3       & 0.045          & 0.058           & 0       & d6       & 0.038          & 0.025           & 0       \tabularnewline\hline
t3       & 0.057          & 0.072           & 0       & v6       & 0.005          & 0.011           & 0       \tabularnewline\hline
q3       & 0.001          & 0.004           & 0.00001 & f6       & 0.006          & 0.012           & 0       \tabularnewline\hline
d3       & 0.028          & 0.038           & 0.00003 & o6       & 0.041          & 0.052           & 0.00001 \tabularnewline\hline
h3       & 0.006          & 0.011           & 0.00046 & e6       & 0.121          & 0.105           & 0.00017 \tabularnewline\hline
k4       & 0.041          & 0.014           & 0       &          &                &                 &        \tabularnewline\hline
\end{tabular}
\end{table}

Table \ref{tab:66-common} shows 66 variables are common as between the first and the current dataset. The existence of common variables between the two datasets is good evidence that a similar phenomenon is operating in relation to the difference between simple and complex words in both datasets.

\begin{table}[H]
\centering
\caption{66 Significant Variables Common to Experiments 1 \& 2 (Bp $< 0.001$)}
\label{tab:66-common}
\begin{tabular}{|l|l|l|l|l|l|}
\hline
a1 & d4 & h2 & l6 & p3 & u4 \tabularnewline\hline
a2 & d6 & h5 & m2 & p6 & u5 \tabularnewline\hline
a3 & e1 & i1 & m5 & r5 & u6 \tabularnewline\hline
a5 & e2 & i3 & m6 & s1 & v5 \tabularnewline\hline
a6 & e5 & i4 & n2 & s2 & v6 \tabularnewline\hline
b1 & e6 & i5 & n4 & s3 & w1 \tabularnewline\hline
b2 & f1 & i6 & n5 & s5 & w2 \tabularnewline\hline
c3 & f3 & j1 & o3 & t1 & w4 \tabularnewline\hline
c5 & g1 & k4 & o4 & t3 & x2 \tabularnewline\hline
c6 & g4 & k5 & o5 & t6 & y2 \tabularnewline\hline
d1 & h1 & l2 & o6 & u1 & y5 \tabularnewline\hline
\end{tabular}
\end{table}

\subsection{Word Rankings}

We create a classifier using the same parameters as in Experiment 1. After deletion of null variables, we have 443 features in the input array. The target variable is a binary value of 1 if the word is in the complex class and 0 if it is in the simple class.  The induction method we use is the Python skikit Learn random forest with n\_estimators=100, max\_depth=None, and min\_samples\_split=5. We use SMOTE on the training data to balance the class sizes. We use 10-fold cross validation, with the scores for each word based on the held out fold. The score cutpoint for classification accuracy is 0.5. 

The accuracy of the classifier is 77\%, (\,baseline = 51\% )\, with 78\% sensitivity and 75\% specificity. The Kappa value is 0.5275 (z = 110.03, p <.0001 – two tailed test). Thus, the classifier is able to efficiently classify the dataset. This indicates that the score given to each word is a reasonable estimate of the word’s complexity. 

One the basis of the classifier score we can rank the words in the sample. Table \ref{tab:word_lengths_exp2} shows the results.

\begin{table}[H]
\footnotesize
\centering
\caption{Lowest and Highest Scoring Words for each Word Length Category - Experiment 2}
\label{tab:word_lengths_exp2}
\begin{tabular}{|l|l|l|l|l|}
\hline
Length      & Word                            & Class      & Score         & Syllables  \tabularnewline\hline
\textbf{2}  & \textbf{bo}                     & \textbf{0} & \textbf{0.03} & \textbf{1} \tabularnewline\hline
\textbf{2}  & \textbf{un}                     & \textbf{0} & \textbf{0.82} & \textbf{1} \tabularnewline\hline
3           & pin                             & 0          & 0.01          & 1          \tabularnewline\hline
3           & amy                             & 0          & 0.78          & 2          \tabularnewline\hline
\textbf{4}  & \textbf{dent}                   & \textbf{0} & \textbf{0.01} & \textbf{1} \tabularnewline\hline
\textbf{4}  & \textbf{unca}                   & \textbf{1} & \textbf{0.89} & \textbf{2} \tabularnewline\hline
5           & stack                           & 0          & 0.01          & 1          \tabularnewline\hline
5           & unset                           & 1          & 0.96          & 2          \tabularnewline\hline
\textbf{6}  & \textbf{leaked}                 & \textbf{0} & \textbf{0.01} & \textbf{1} \tabularnewline\hline
\textbf{6}  & \textbf{uncase}                 & \textbf{1} & \textbf{0.96} & \textbf{2} \tabularnewline\hline
7           & showing                         & 0          & 0.02          & 2          \tabularnewline\hline
7           & uncanny                         & 1          & 0.96          & 3          \tabularnewline\hline
\textbf{8}  & \textbf{shipping}               & \textbf{0} & \textbf{0.02} & \textbf{2} \tabularnewline\hline
\textbf{8}  & \textbf{unwisely}               & \textbf{1} & \textbf{0.98} & \textbf{3} \tabularnewline\hline
9           & newspaper                       & 0          & 0.09          & 3          \tabularnewline\hline
9           & advisably                       & 1          & 0.99          & 4          \tabularnewline\hline
\textbf{10} & \textbf{storyboard}             & \textbf{0} & \textbf{0.08} & \textbf{3} \tabularnewline\hline
\textbf{10} & \textbf{unsuitable}             & \textbf{1} & \textbf{1}    & \textbf{4} \tabularnewline\hline
11          & spokeswoman                     & 0          & 0.07          & 3          \tabularnewline\hline
11          & unalterable                     & 1          & 1             & 5          \tabularnewline\hline
\textbf{12} & \textbf{neighborhood}           & \textbf{0} & \textbf{0.16} & \textbf{3} \tabularnewline\hline
\textbf{12} & \textbf{discerningly}           & \textbf{1} & \textbf{1}    & \textbf{4} \tabularnewline\hline
13          & southwestward                   & 0          & 0.18          & 3          \tabularnewline\hline
13          & fantastically                   & 1          & 1             & 4          \tabularnewline\hline
\textbf{14} & \textbf{newspaperwoman}         & \textbf{0} & \textbf{0.17} & \textbf{5} \tabularnewline\hline
\textbf{14} & \textbf{unfaithfulness}         & \textbf{1} & \textbf{1}    & \textbf{4} \tabularnewline\hline
15          & environmentally                 & 1          & 0.21          & 6          \tabularnewline\hline
15          & intolerableness                 & 1          & 1             & 7          \tabularnewline\hline
\textbf{16} & \textbf{environmentalism}       & \textbf{1} & \textbf{0.15} & \textbf{7} \tabularnewline\hline
\textbf{16} & \textbf{discriminatively}       & \textbf{1} & \textbf{1}    & \textbf{6} \tabularnewline\hline
17          & anesthesiologists               & 1          & 0.48          & 6          \tabularnewline\hline
17          & intellectualising               & 1          & 1             & 7          \tabularnewline\hline
\textbf{18} & \textbf{electrocardiograph}     & \textbf{0} & \textbf{0.26} & \textbf{7} \tabularnewline\hline
\textbf{18} & \textbf{comprehensibleness}     & \textbf{1} & \textbf{1}    & \textbf{6} \tabularnewline\hline
19          & electrocardiography             & 0          & 0.3           & 8          \tabularnewline\hline
19          & intellectualization             & 1          & 1             & 8          \tabularnewline\hline
\textbf{20} & \textbf{hypercholesterolemia}   & \textbf{1} & \textbf{0.71} & \textbf{9} \tabularnewline\hline
\textbf{20} & \textbf{incomprehensibleness}   & \textbf{1} & \textbf{1}    & \textbf{7} \tabularnewline\hline
21          & electroencephalograph           & 1          & 0.77          & 5          \tabularnewline\hline
21          & indistinguishableness           & 1          & 0.91          & 7          \tabularnewline\hline
\textbf{22} & \textbf{deinstitutionalization} & \textbf{1} & \textbf{0.9}  & \textbf{8} \tabularnewline\hline
\textbf{22} & \textbf{counterrevolutionaries} & \textbf{1} & \textbf{0.94} & \textbf{8} \tabularnewline\hline
\end{tabular}
\end{table}

The number of word-length pairs in which the number of syllables for the lower scoring word is greater than that for the higher scoring word is 4, while the figure is 13 for the pairs where the higher scoring word is has a greater number of syllables. In short, controlling for word length, the number of syllables in complex words is greater than in simple words. This is essentially what we observed in the initial dataset.

\section{Experiment 3}

In this experiment, we create classifiers using as features those variables we have found to be significant above. We create a third dataset to use for this experiment. The idea is that if the variables we have identified as significant above are associated with LC, we should be able to use them to classify a dataset consisting of a high LC class and a low LC class significantly better than what we would expect by chance alone.

The data for the simple word group comes from the The Children and Young People’s Books Lexicon
(CYP-LEX) \parencite{korochkina2023children}. We take the 10000 most common word lemmas that appear in books written for 7-9 year olds. The data for the complex dataset comes from the COCA Academic 20000 corpus. These are word lemmas that are commonly used in academic texts. We remove the hyphenated words and rank the remainder by frequency in the general COCA corpus and take the least common 10000 words. We use the same filtering methods used in Experiments 1 and 2.

\begin{table}[H]
\centering
\caption{Word Filtering - Experiment 3}
\label{tab:Word Filtering - Experiment 3}
\begin{tabular}{|l|l|l|}
\hline
                                 & Simple & Complex \tabularnewline\hline
Total Words                      & 10000  & 10000   \tabularnewline\hline
Cleaned Words                    & 9968   & 9990    \tabularnewline\hline
Within Class Duplicates Removed  & 9968   & 9688    \tabularnewline\hline
Between Class Common Words Removed & 8754   & 8474   \tabularnewline\hline
\end{tabular}
\end{table}

To select variables for the classifiers, we use the results from Experiments 1 and 2. We wish to see if the significant variables we have isolated in these experiments are able to correctly classify complex and simple words from a separate dataset. We use the 66 significant variables we have identified in the first two experiments. We use these for two reasons. The first is that these were all significant at the Bp $< .001$ level in both Experiments 1 and 2. The idea here is that we want only those variables that are likely to generalise to other datasets, so a variable that has been shown to be significant in two separate datasets is more likely to generalise to a third. The second reason for using the common 66 variables is that they are all focused on the first 6 positions. This is important because this ensures we are looking at letter positions on the basis of factors other than length. Allowing length to influence variable selection would provide a signal that biases the classifier due to complex words generally being longer than simple words.
 
 Using the same classifier parameters as in Experiments 1 and 2, we create a classifier using the 66 significant variables. As a comparison, we draw 66 random variables from the 403 non-null variables to create a second classifier and we create a third classifier using all 403 non-null variables in the dataset. Results for the three classifiers are in Table \ref{tab:results_for_three_classifiers}. Note that all Kappa values are significant at the p$<.0001$ level, and the baseline accuracy = 50\%

\begin{table}[H]
\centering
\caption{Accuracy, Sensitivity and Specificity for 3 Classifiers}
\label{tab:results_for_three_classifiers}
\begin{tabular}{|l|l|l|l|l|}
\hline
Variables in Classifier  & Acc'\% & Sens'\% & Spec'\% & Kappa  \tabularnewline\hline
66 Significant Variables & 70         & 69            & 72            & 0.4092 \tabularnewline\hline
66 Random Variables      & 65         & 52            & 79            & 0.3065 \tabularnewline\hline
All 403 variables        & 78         & 75            & 81            & 0.5602 \tabularnewline\hline
\end{tabular}
\end{table}

The 66 variables we have identified in Experiments 1 and 2 as significant in the first 6 letter positions are  sufficient to classify the dataset at 70\% accuracy in comparison to 65\% accuracy using 66 random variables. Thus, the same variables we identified in Experiments 1 and 2 can be used to classify words in the third dataset beyond what we would achieve by chance. Adding all 403 non-null variables to the classifier results in an accuracy of 78\%.

\section{Application}

We have used three distinct datasets to design and test several classifiers. In this section we use the results of the previous classifiers to compile a set of high LC words and a set of low LC simple words. We then use these to create a classifier which we use to score a dictionary 128511 English words.

The basic approach here is to use the upper and lower extremes of the scores for the words generated in each Experiment above as the basis for the simple and complex classes. Note that the word list used from Experiment 3 is that for the classifier with the full cohort of 403 variables, as it is more accurate than the classifiers with 66 variables.

To create the simple class, we take the scores for the classifiers in Experiments 1, 2 and 3 that are below 0.3, which gives a simple word class of 20628. For the complex class, we take the words that score above 0.7, which provides a total of 27295 words.  Removing duplicates within and common words between classes results in a dataset of 13565 simple words and 21999 complex words. Removing null variables gives us a feature set of 443 variables.

Using the same parameters for the classifier as above, the accuracy is 97\%, with 99\% sensitivity and 95\% specificity (Kappa = 0.9430, Z = 177.87, p$<.0001$). 

The corpus to be scored consists of 168065 words of the Hunspell English (US) spellchecker dictionary \parencite{hunspell_dictionary}. We clean the data by removing words with non-alphabetic characters and removing individual letters and duplicates. This leaves a sample of 128511 words, including capital cities and personal names. Descriptive statistics for the scored corpus are displayed in Table \ref{tab:Statistics-for-Scored-Words}.

\begin{table}[H]
\centering
\caption{Statistics for Scored Words}
\label{tab:Statistics-for-Scored-Words}
\begin{tabular}{|l|l|}
\hline
Average  & 0.631  \tabularnewline\hline
Stdev    & 0.38   \tabularnewline\hline
Median   & 0.81   \tabularnewline\hline
Min      & 0      \tabularnewline\hline
Max      & 1      \tabularnewline\hline
Kurtosis & -1.247 \tabularnewline\hline
Skew     & -0.63  \tabularnewline\hline
\end{tabular}
\end{table}

To contextualise these figures, we score a sample of 2295 words used for the CEFR-J ESL program \parencite{CEFR-J}. These `B2 level' words are indicative of an `independent' level of English, which is below the CEFR-J `proficient' level \parencite{masashi2012development}. The average of the classifier scores for these words is 0.655 which is slightly higher than the average dictionary score of 0.631. Thus, the average score of the dictionary is close to the complexity of words we might find being used in an intermediate Japanese ESL class. To further contextualise the dictionary average score, Table \ref{tab:words_in_the_average_range} shows a sample of words from the dictionary in the range of  0.631.

\begin{table}[H]
\centering
\caption{Words in the Dictionary Average Range}
\label{tab:words_in_the_average_range}
\begin{tabular}{|l|l|}
\hline
Word      & Score \tabularnewline\hline
sprockets & 0.630  \tabularnewline\hline
frazzling & 0.630  \tabularnewline\hline
guiltiest & 0.631 \tabularnewline\hline
slenderer & 0.631 \tabularnewline\hline
ectotherm & 0.631 \tabularnewline\hline
\textbf{fireballs} & \textbf{0.631}  \tabularnewline\hline
heartiest & 0.631 \tabularnewline\hline
signalman & 0.631 \tabularnewline\hline
flambeaux & 0.631 \tabularnewline\hline
responsum & 0.631 \tabularnewline\hline
flashiest & 0.631 \tabularnewline\hline
\end{tabular}
\end{table}

Table \ref{tab:highest_and_lowest-4} shows the highest and lowest scoring words for word length categories 2 -22.

\begin{table}[H]
\footnotesize
\centering
\caption{Highest and Lowest Scores for Word Length Categories - Experiment 4}
\label{tab:highest_and_lowest-4}
\begin{tabular}{|l|l|l|l|}
\hline
Length      & Word                            & Score          & Syllables  \tabularnewline\hline
\textbf{2}  & \textbf{al}                     & \textbf{0}     & \textbf{1} \tabularnewline\hline
\textbf{2}  & \textbf{un}                     & \textbf{0.917} & \textbf{1} \tabularnewline\hline
3           & aim                             & 0              & 1          \tabularnewline\hline
3           & una                             & 0.933          & 2          \tabularnewline\hline
\textbf{4}  & \textbf{ahoy}                   & \textbf{0}     & \textbf{2} \tabularnewline\hline
\textbf{4}  & \textbf{undo}                   & \textbf{0.977} & \textbf{2} \tabularnewline\hline
5           & awoke                           & 0              & 2          \tabularnewline\hline
5           & unmet                           & 0.998          & 2          \tabularnewline\hline
\textbf{6}  & \textbf{baddie}                 & \textbf{0}     & \textbf{2} \tabularnewline\hline
\textbf{6}  & \textbf{unsure}                 & \textbf{0.995} & \textbf{2} \tabularnewline\hline
7           & babbled                         & 0              & 2          \tabularnewline\hline
7           & unmoral                         & 1              & 3          \tabularnewline\hline
\textbf{8}  & \textbf{batching}               & \textbf{0}     & \textbf{2} \tabularnewline\hline
\textbf{8}  & \textbf{yemenite}               & \textbf{1}     & \textbf{3} \tabularnewline\hline
9           & glasswork                       & 0.009          & 2          \tabularnewline\hline
9           & zealously                       & 1              & 3          \tabularnewline\hline
\textbf{10} & \textbf{scoreboard}             & \textbf{0.034} & \textbf{2} \tabularnewline\hline
\textbf{10} & \textbf{vulnerable}             & \textbf{1}     & \textbf{4} \tabularnewline\hline
11          & craftswoman                     & 0.035          & 3          \tabularnewline\hline
11          & zealousness                     & 1              & 3          \tabularnewline\hline
\textbf{12} & \textbf{warehouseman}           & \textbf{0.045} & \textbf{3} \tabularnewline\hline
\textbf{12} & \textbf{zoochemistry}           & \textbf{1}     & \textbf{4} \tabularnewline\hline
13          & grandfathered                   & 0.073          & 3          \tabularnewline\hline
13          & vulnerability                   & 1              & 6          \tabularnewline\hline
\textbf{14} & \textbf{newspaperwoman}         & \textbf{0.092} & \textbf{5} \tabularnewline\hline
\textbf{14} & \textbf{victoriousness}         & \textbf{1}     & \textbf{5} \tabularnewline\hline
15          & southwestwardly                 & 0.202          & 4          \tabularnewline\hline
15          & vulnerabilities                 & 1              & 6          \tabularnewline\hline
\textbf{16} & \textbf{straightforwards}       & \textbf{0.338} & \textbf{3} \tabularnewline\hline
\textbf{16} & \textbf{unresponsiveness}       & \textbf{1}     & \textbf{5} \tabularnewline\hline
17          & straightforwardly               & 0.328          & 4          \tabularnewline\hline
17          & unpretentiousness               & 1              & 5          \tabularnewline\hline
\textbf{18} & \textbf{electrocardiograph}     & \textbf{0.233} & \textbf{7} \tabularnewline\hline
\textbf{18} & \textbf{unconstitutionally}     & \textbf{1}     & \textbf{7} \tabularnewline\hline
19          & electrocardiography             & 0.223          & 8          \tabularnewline\hline
19          & unpretentiousnesses             & 1              & 6          \tabularnewline\hline
\textbf{20} & \textbf{electrocardiographic}   & \textbf{0.233} & \textbf{8} \tabularnewline\hline
\textbf{20} & \textbf{overenthusiastically}   & \textbf{1}     & \textbf{8} \tabularnewline\hline
21          & otorhinolaryngologist           & 0.833          & 9          \tabularnewline\hline
21          & uncommunicativenesses           & 1              & 8          \tabularnewline\hline
\textbf{22} & \textbf{otorhinolaryngologists} & \textbf{0.833} & \textbf{9} \tabularnewline\hline
\textbf{22} & \textbf{deinstitutionalization} & \textbf{1}     & \textbf{9} \tabularnewline\hline
\end{tabular}
\end{table}

In order to test the efficacy of the scoring, we need a list of words scored by some means on complexity. As a test of the efficacy of the scoring, we take the sample of words from four graduated levels of the CEFR-J program cited above. These 7799 words are rated in four categories from A1 - A2 (Basic) to B1 - B2 (Independent). We remove words with any non-alphabetic characters and single letter words. We also remove words that appear in more than one category due to different function, such as ‘ache’ as a noun (B1) and as a verb (B2). Scoring data for the words extracted using this procedure is presented in Table \ref{tab:scores_for_cefr}

\begin{table}[H]
\centering
\caption{Scores for CEFR-J Levels A1 - B2}
\label{tab:scores_for_cefr}
\begin{tabular}{|l|l|l|l|l|}
\hline
Level         & A1    & A2    & B1    & B2    \tabularnewline\hline
No. Words     & 799   & 971   & 1916  & 2295  \tabularnewline\hline
Average Score & 0.164 & 0.414 & 0.576 & 0.655 \tabularnewline\hline
Stdev         & 0.298 & 0.402 & 0.417 & 0.416 \tabularnewline\hline
\end{tabular}
\end{table}

As we might expect, scores have a graduated increasing average across the four categories. However, there is a high degree of variation within each category. The average plus one standard deviation of the first category  = 0.164 + 0.298 = 0.463, which is above the average of the second category, and the same pattern occurs for the remaining 3 categories. Thus, for all categories, there are words that score in the average range for the given category, but which are higher than the scores of many of the words in the higher category scoring in the average range. A similar phenomenon occurs in the opposite direction, with the average – 1 standard deviation for higher levels being lower than the average for the next lowest level. In short, there is a high level of overlap in the scores of the four categories.

If we consider the two extreme classes – A1 and B2 – we find that there are 8 words in A1 that score 1 (the extreme high end of the classifier spectrum), and 284 that score 0 (the extreme low end of the classifier spectrum) in B2. This is not an ideal situation if the categories are intended to be indicative of complexity. In order to examine what is happening here, we look at two significant errors: the word ‘celebration’ in A1, which scores 1, but should score significantly lower given its inclusion in the basic English category, and the word ‘badge’ in B2 which scores 0, but should score significantly higher given its inclusion in the independent category. Table \ref{tab:celebration_badge} shows the words in the range of alphabetically similarly structured words for both of these words.

\begin{table}[H]
\centering
\caption{Scores for Words of Similar Structure to `Celebration' and `Badge'}
\label{tab:celebration_badge}
\begin{tabular}{|l|l||l|l|}
\hline
Word                 & Score      & Word           & Score      \tabularnewline\hline
celandine            & 0.976      & badajoz        & 0.414      \tabularnewline\hline
celanese             & 0.836      & badalona       & 0.835      \tabularnewline\hline
celeb                & 0.091      & badder         & 0.001      \tabularnewline\hline
celebes              & 0.174      & badderlocks    & 0.874      \tabularnewline\hline
celebrant            & 0.977      & baddest        & 0.073      \tabularnewline\hline
celebrants           & 0.988      & baddie         & 0          \tabularnewline\hline
celebrate            & 0.964      & baddies        & 0          \tabularnewline\hline
celebrated           & 0.965      & baddy          & 0          \tabularnewline\hline
celebrates           & 0.974      & bade           & 0          \tabularnewline\hline
celebrating          & 0.998      & baden          & 0.013      \tabularnewline\hline
\textbf{celebration} & \textbf{1} & \textbf{badge} & \textbf{0} \tabularnewline\hline
celebrations         & 1          & badged         & 0          \tabularnewline\hline
celebrator           & 0.941      & badger         & 0          \tabularnewline\hline
celebrators          & 0.957      & badgered       & 0.073      \tabularnewline\hline
celebratory          & 0.957      & badgering      & 0.523      \tabularnewline\hline
celebrities          & 1          & badgers        & 0          \tabularnewline\hline
celebrity            & 0.979      & badges         & 0.002      \tabularnewline\hline
celebrityhood        & 0.969      & badinage       & 0.481      \tabularnewline\hline
celebs               & 0.096      & badland        & 0.492      \tabularnewline\hline
celeriac             & 0.93       & badlands       & 0.464      \tabularnewline\hline
celeriacs            & 0.953      & badly          & 0         \tabularnewline\hline
\end{tabular}
\end{table}

Celebration scores higher than the longer ‘celebrityhood’ (0.969), and higher than the equally long ‘celebratory’ (0.957). The high score is not directly due to the initial letters as celeb, celebes, and celebs score quite low.   However, if we look at 11 letter words with ‘ation’ at the 8th position, the average score is 0.9944, STDEV = 0.0121, n = 396. Indicative words include inscription,
lionization, orientation, reinvention and indentation. Thus, we conclude that the high score is valid and is due to the word ending. Its inclusion in A1 may be due to the CEFR-J requirements.

‘Badge’, scores lower than the equally long ‘baden’ (0.013), and the same as the longer ‘badged’,
‘badger’, ‘baddie’ and ‘baddies’. Thus, the score seems appropriate for the structure of the word. Furthermore, the word appears in the training data in the 10,000 most frequent words from the 74,286 Brysbaert and New \parencite{brysbaert2009moving} frequency norms, which indicates that it is a high frequency (low complexity) word. Thus, we conclude that the score for badge is appropriate, and that the inclusion of the word in B2 may be due to the particular requirements of the CEFR program. 

\section{Conclusion}

In this paper, we found that using letter positional probabilities enables us to capture a form of word complexity. In Experiments 1 and 2, we found, using two different datasets, 66 common letter positional probabilities significant at the Bp $<.001$ level. In both datasets, we were able to classify the words at a high level of accuracy using only letter positional probabilities. In Experiment 3, using a third dataset, we used the 66 common variables identified in Experiments 1 and 2 in the first 6 letter positions to create a classifier that was more accurate than 66 variables chosen at random, although neither of the 66-variable classifiers was as accurate at the classifier created with the third dataset using all 443 letter positional probabilities. Finally, we combined the words at the extreme upper and lower scores of the classifiers created with the full cohort of available variables in Experiments 1-3 to create a classifier with a 97\% accuracy. We used this to score a dictionary of 128511 words. We found that the scores were able to score four levels of a sample of CEFR-J words with average scores that reflect the four levels of complexity. We found that the scores had a high level of variation within each class, but an analysis of two examples of the errors indicated that the scores for those examples were explicable in terms of the nature of the words involved.

Importantly, we have found that there are differences between high and low LC words in terms of the vowel/consonant distribution. Vowels are significant in all letter positions but 2 and 3 for high LC words, while the reverse is true for low LC words. We also made the observation that, with words of the same length, more complex words have more syllables.

Overall, the study shows that letter positional probabilities are an effective way of analysing word complexity beyond the standard lexical, psycholinguistic and morphological methods.

\medskip

\raggedright{\textbf{Data Availability Statement:}}
The data used in this study are available at Figshare (https://figshare.com/),
https://doi.org/10.6084/m9.figshare.25591131.v1.\\
\medskip
\textbf{Research Funding Statement:}
This research was supported by an Australian Government RTP Scholarship.\\
\medskip
\textbf{Disclosure Statement:}

The author reports there are no known competing interests to declare.

\medskip

\printbibliography
\end{document}